\pdfoutput=1
\documentclass[11pt,a4paper]{article}

\usepackage[final]{acl}

\usepackage{times}
\usepackage{latexsym}

\usepackage[T1]{fontenc}
\usepackage[utf8]{inputenc}

\usepackage{microtype}

\usepackage{inconsolata}

\usepackage{graphicx}

\usepackage[most]{tcolorbox}
\tcbuselibrary{breakable, listingsutf8}
\title{Agentic LLMs for Question Answering over Tabular Data}

\author{
  Rishit Tyagi* \\
  {\normalsize \texttt{tyagirishit21@gmail.com}} \\
  \And
  Mohit Gupta* \\
  {\normalsize \texttt{mohit.gupta2002@gmail.com}} \\
  \And
  Rahul Bouri* \\
  {\normalsize \texttt{rahulbouri16@gmail.com}} \\
}
\begin{document}
\maketitle

\renewcommand{\thefootnote}{\fnsymbol{footnote}}
\footnotetext[1]{These authors contributed equally to this work.}

\begin{abstract}
Question Answering over Tabular Data (Table QA) presents unique challenges due to the diverse structure, size, and data types of real-world tables. The SemEval 2025 Task 8 (DataBench) introduced a benchmark composed of large-scale, domain-diverse datasets to evaluate the ability of models to accurately answer structured queries. We propose a Natural Language to SQL (NL-to-SQL) approach leveraging large language models (LLMs) such as GPT-4o, GPT-4o-mini, and DeepSeek v2:16b to generate SQL queries dynamically. Our system follows a multi-stage pipeline involving example selection, SQL query generation, answer extraction, verification, and iterative refinement. Experiments demonstrate the effectiveness of our approach, achieving 70.5\% accuracy on DataBench QA and 71.6\% on DataBench Lite QA, significantly surpassing baseline scores of 26\% and 27\% respectively. This paper details our methodology, experimental results, and alternative approaches, providing insights into the strengths and limitations of LLM-driven Table QA.
\end{abstract}

\section{Introduction}
Question Answering over Tabular Data (Table QA) is a fundamental problem in natural language processing (NLP) that aims to retrieve structured information from tables given natural language queries. This task is crucial for making structured data more accessible, enabling users to interact with databases, spreadsheets, and structured documents without requiring expertise in SQL or database querying. However, Table QA presents unique challenges compared to traditional open-domain QA, as it requires models to understand schema structures, perform logical reasoning over tabular relationships, and generate precise queries that extract relevant data \cite{solimansurvey}.
\newline Early approaches to Table QA relied on rule-based systems \cite{4312923}, manually designed templates, and retrieval-augmented generation (RAG) \cite{pan2022end}. While effective for constrained domains, these methods struggle with scalability, particularly when handling large and diverse tables with complex relationships. More recent advances employ neural models for end-to-end Table QA, including methods that directly generate SQL queries from natural language inputs. Large language models (LLMs) have demonstrated impressive capabilities in this space, enabling the dynamic generation of SQL queries without requiring predefined schemas or manually curated rules \cite{baig2022natural}.\newline Despite these advancements, several challenges remain. LLM-generated SQL queries often suffer from structural errors, incorrect column selections, and ambiguous reasoning over tabular data. Additionally, extracting precise answers from retrieved SQL results requires careful post-processing, as LLMs may misinterpret numerical values, categories, or required formats. To address these issues, verification and refinement steps are necessary to ensure query correctness and answer reliability.
\newline In this work, we present an LLM-driven Natural Language to SQL (NL-to-SQL) pipeline that dynamically translates user questions into SQL queries, retrieves structured data, and refines the final output to maximize answer accuracy. Our approach integrates multiple stages, including example selection, query generation, answer extraction, and verification. By leveraging LLMs for both SQL generation and answer refinement, we aim to improve robustness across diverse table structures and query types.
We apply our system to the SemEval 2025 Task 8 (DataBench), a benchmark designed to evaluate Table QA over real-world datasets.  By refining query formulation and integrating a verification mechanism, our system significantly outperforms baselines. In this paper, we detail our system architecture, performance evaluation, and key insights derived from our experiments.

\section{Related Work}

The task of Question-Answering on Tabular Data (Table QA) involves extracting precise, grounded answers from structured data based on natural language queries. Various methods have been explored to tackle this problem, yet challenges such as data sparsity, feature heterogeneity, context-based interconnections, and order invariance remain significant \cite{fang2024largelanguagemodelsllmstabular}. Previous research in Table QA has introduced generative, extractive, and retriever-reader-based methods, each addressing different aspects of reasoning and information retrieval \cite{jin2022surveytablequestionanswering}.

Generative models, such as those introduced in \cite{pasupat2015compositionalsemanticparsingsemistructured} generate answers directly instead of producing logical forms. Extractive methods, in contrast, select spans of text from tables rather than generating them, relying on effective table cell representations to capture only relevant information. Structure-aware approaches like TAPAS \cite{Herzig_2020} incorporate row/column embeddings to encode positional information, while models like TableFormer \cite{gupta2022tableformer} introduce attention bias techniques to enhance table reasoning. 
While both generative and extractive models excel at handling simple queries, their performance deteriorates on reasoning-based queries that require logical inference and multi-hop reasoning. \cite{jin2022surveytablequestionanswering}

Among these, NL-to-SQL has emerged as a powerful approach \cite{mohammadjafari2025naturallanguagesqlreview}, translating natural language queries into structured SQL statements for efficient information retrieval. Traditional NL-to-SQL models followed encoder-decoder-based architectures suited for structured databases, but real-world applications often involve semi-structured and free-form tables, necessitating alternative techniques \cite{hong2025nextgenerationdatabaseinterfacessurvey}. The advent of large language models (LLMs) has brought a paradigm shift to NL-to-SQL. Unlike traditional chat-based completion models, reasoning-driven LLMs excel at understanding complex question intent, handling multi-step logical reasoning, and adapting to diverse database schemas with minimal training. Researchers have also experimented with prompt engineering and fine-tuning to improve SQL query generation efficiency. Chain-of-Thought (CoT) prompting enables LLMs to break down complex queries step by step, further enhancing reasoning capabilities.

LLM-based approaches have demonstrated superior evaluation metrics on benchmark datasets such as SPIDER \cite{yu2019spiderlargescalehumanlabeleddataset}, surpassing traditional models. As research progresses, improvements in model size, reasoning capabilities, and dataset quality are expected to drive further performance gains, solidifying LLMs as the dominant paradigm for Table QA.

\section{Task Description}
The SemEval 2025 Task 8 \cite{oses-etal-2024-databench}, known as DataBench, is designed to evaluate systems that answer questions using real-world tabular datasets. The challenge comprises two subtasks:

\begin{enumerate}
    \item \textbf{DataBench QA}: Participants are provided with entire datasets and corresponding questions, requiring systems to extract answers from potentially large and complex tables.
    \item \textbf{DataBench Lite QA}: This subtask involves answering questions using a sampled version of each dataset, limited to a maximum of 20 rows, focusing on models' ability to handle smaller data contexts.
\end{enumerate}
The DataBench benchmark encompasses 65 diverse real-world datasets, each accompanied by multiple questions. These datasets span various domains and exhibit a wide range of data types and table sizes, challenging systems to adapt to different structures and content. The questions associated with these datasets are designed to elicit various response types, including numerical values, categorical data, boolean judgments, and lists. This diversity necessitates that participating systems possess robust capabilities in understanding and processing heterogeneous tabular data to generate accurate and contextually appropriate answers.

\section{Methodology}
Our approach leverages a Natural Language to SQL (NL-to-SQL) agent to generate structured queries that retrieve relevant information from tabular data. The system follows a multi-stage pipeline consisting of example selection, SQL query generation, answer extraction, verification, and iterative refinement. This framework ensures high accuracy and adaptability across datasets with varying structures and question complexities. Alternative methods such as rule-based retrieval and RAG were tested but proved less effective, especially when handling large-scale tabular reasoning tasks.
\newline Additionally, we experimented with multiple models on the same pipeline to optimize performance, including GPT-4o, GPT-4o-mini, and DeepSeek v2:16b. Our system follows the following five-stage pipeline.

\subsection{Example Selection}
To improve SQL query generation, we curated a set of 25 example question-query pairs, where each question is a natural language input and each query is the corresponding SQL output, covering diverse question patterns such as filtering, aggregation, grouping, sorting, joins, subqueries, and conditional retrievals \cite{nan2023enhancing}. These examples were designed to represent various structural and semantic variations commonly found in tabular data questions. Given an input question, we computed cosine similarity with these pre-defined examples and selected the two most relevant question-query pairs to provide as context.
\newline
To determine the most relevant examples for a given input question, we utilized an embedding-based similarity approach. First, we generated vector representations (embeddings) for all example questions using text-embedding-ada-002 and stored it in a ChromaDB vector database collection. When a new question was received, we computed its embedding using the same model and calculated the cosine similarity between the embedding of the input question and those of the pre-defined examples in the collection. The two most similar examples were selected as context for SQL generation, ensuring that the model received relevant references closely matching the structure and intent of the input question. This approach helped guide the model in producing more precise and contextually appropriate queries, improving the accuracy of our system.

\subsection{SQL Query Generation}
To generate accurate SQL queries, we provided the LLM with a structured prompt that included the table schema, specifying column names and data types, along with a few sample rows to offer contextual grounding \cite{wu2024cooperativesqlgenerationsegmented}. Additionally, we incorporated the two most relevant example natural language question-sql query pairs, selected based on cosine similarity, to guide the model in producing syntactically and semantically appropriate SQL statements. To enforce query validity, explicit SQL syntax constraints were included in the prompt, ensuring that the generated queries adhered to the expected format. The primary objective of the initial retrieval step was to extract relevant table rows rather than compute direct answers, allowing for a structured query execution process. These generated SQL queries were then executed on the SQLite database containing the DataBench datasets, retrieving the necessary rows, which were subsequently processed in later stages to derive the final answers.

\subsection{Answer Extraction and Formatting}
Once rows were retrieved via SQL, a secondary prompt analyzed and extracted the data to derive the final answer, ensuring compliance with expected data types (e.g., ordered lists, numerical outputs). This step was crucial for refining the raw SQL outputs into the structured response format required by the task.
To enhance the reasoning capability of the model, we employed Chain-of-Thought (CoT) prompting \cite{liu2023dividepromptchainthought}, which allowed the LLM to break down the answer derivation process into logical steps. The model was instructed to analyze the retrieved rows, extract only the necessary values, and format them correctly based on the expected answer type.
\newline The prompt provided structured guidelines for different expected output types, including boolean values, categorical selections, numerical computations, and list-based answers. By leveraging structured CoT reasoning, the LLM could systematically evaluate the retrieved data, determine the most relevant cell or computed value, and return a precise answer aligned with the question intent.

\subsection{Answer Verification}
A verification step using an additional GPT-4o prompt classified responses based on two key criteria: format validity and relevance to the given question \cite{wang2024toolassistedagentsqlinspection}. The system was designed to assess whether an answer adhered to expected formats—such as Boolean values, numerical values, dates, or categorical lists—without performing fact-checking. If an answer deviated from these predefined formats or was entirely unrelated to the question, it was flagged for reprocessing. To ensure robustness, borderline cases were defaulted to acceptance unless a clear and significant violation of format or relevance was detected. This step minimized incorrect responses by identifying cases where the initial SQL query retrieved extraneous or irrelevant information, ensuring that only properly structured and contextually appropriate answers progressed to the final stage.

\subsection{Answer Reprocessing}
If a response was flagged for reprocessing, the system adapted its approach to improve answer precision. The examples used in SQL generation were updated to prioritize queries that retrieved specific values rather than entire rows, ensuring a more targeted extraction of relevant information. Additionally, the SQL generation prompt was refined to directly extract the exact answer values from the dataset, minimizing unnecessary retrieval of irrelevant data. A final formatting step enforced consistency with the expected output type, whether numerical, categorical, or list-based, aligning responses with the required structure. The entire query generation and execution pipeline was re-executed after these refinements were made on queries that were flagged for reprocessing. This iterative refinement process enhanced overall answer correctness and significantly reduced extraneous outputs. 
\newline After reprocessing, the outputs from the newly refined queries were combined with the outputs from the original successful (approved) queries. If a query was flagged and reprocessed, its new result replaced the earlier one. If a query was not flagged, its original output was retained. This way, the final set of results included the best available answer for each query—either from the initial run (if it was already correct) or from the reprocessed run (if it had been improved). The system ensured there were no duplicates and that each query had exactly one final, verified result in the merged output.

\section{Results}
To evaluate the effectiveness of our Natural Language to SQL (NL-to-SQL) query agent, we conducted experiments on two benchmark datasets - Databench and Databench-Lite, using three Large Language Models (LLMs): GPT-4o, GPT-4o-mini, and deepseek-v2:16b \cite{DBLP:journals/corr/abs-2405-04434}. Additionally, we compared our results to the provided baseline model to establish a reference point for performance \cite{sinha2024evaluating}. \newline GPT-4o consistently demonstrated superior performance compared to other models. Its ability to accurately interpret and break down user queries played a crucial role in generating results in the expected format. This allowed for more precise and contextually relevant outputs, particularly in complex scenarios. GPT-4o-mini also exhibited many of the advantages of GPT-4o, particularly in question understanding and structured output generation, though to a lesser extent due to its smaller model size and optimization for cost-effective applications. DeepSeek v2:16b showed some capability in processing structured queries but lacked the same level of precision, adaptability, and ability to follow prompt instructions. 
\newline For evaluation, the organizers rank the system
based on accuracy of the answers on the question answering task.  Our approach ranked 10th on the Databench task proprietary model leaderboard and 9th on the Databench-Lite task proprietary model leaderboard, demonstrating the capabilities of our system. Table 1 presents the accuracy of our models compared to the baseline.
\newline Our NL-to-SQL pipeline, combined with LLM capabilities, resulted in substantial improvements over the baseline across all models. The enhancements introduced in our approach directly contributed to these improvements:
\begin{enumerate}
\item  Embedding-based example selection improved query contextualization and standardization, leading to more accurate SQL generation and a higher success rate for complex queries.
\item  Chain-of-Thought (CoT) reasoning significantly reduced errors in answer extraction, particularly in handling numerical computations, categorical selections, and ordered lists.
\item Answer verification and iterative reprocessing helped eliminate hallucinated SQL queries and irrelevant outputs, ensuring greater response reliability.
\end{enumerate}
These methodological improvements enabled GPT-4o to achieve the highest accuracy (70.50\% on Databench and 71.65\% on Databench-Lite), clearly outperforming the baseline performance. Even deepseek-v2:16b, despite its lower overall accuracy, showed a significant improvement over the baseline, demonstrating the effectiveness of our multi-stage NL-to-SQL framework.

\begin{table}[t]
    \centering
    \captionsetup{font=normalsize}

    \begin{minipage}{\linewidth}
        \centering
        \begin{tabular}{cccccc}
            \hline
            Model & Databench & Databench-lite \\
            \hline
            Baseline & 26.00 & 27.00 \\
            
            GPT-4o-mini & 60.34 & 61.49 \\
           
            GPT-4o & \textbf{70.50} & \textbf{71.65} \\
            
            deepseek-v2:16b & 39.68 & 45.78 \\
            \hline
        \end{tabular}
        \caption{Accuracy Score Comparison across models}
        \label{tab:first}
    \end{minipage}
\end{table}

\section{Conclusion}
This study advances NL-to-SQL translation by developing a multi-stage LLM-driven agentic pipeline that significantly improves query accuracy, explanability and query consistency compared to previous models. Our approach integrates example selection, Chain-of-Thought (CoT) reasoning, and answer verification, which collectively enhance SQL generation by improving query structure consistency, reducing ambiguity, and refining output accuracy. GPT-4o achieved the highest accuracy across both benchmark datasets, demonstrating the effectiveness of this structured approach over baseline methods.
\newline By leveraging context-aware question selection and structured reasoning, the system effectively mitigates common NL-to-SQL challenges, such as misinterpretation of input question intent and incorrect SQL syntax. The incorporation of an iterative refinement process further ensures robust query generation, reducing errors, and enhancing the overall accuracy. 
\newline The results on smaller datasets like Databench-Lite illustrate the effectiveness of our pipeline in generating accurate and well-structured SQL queries for constrained datasets. Furthermore, the consistent performance observed on larger datasets demonstrates the scalability of our framework, making it well-suited for enterprise applications that require high reliability and adaptability across diverse corporate databases. This robustness ensures that organizations can leverage our approach for complex query generation at scale, improving efficiency and decision-making processes.
\newline These findings reinforce the potential of LLM-driven NL-to-SQL systems as a scalable and efficient solution for automating database interactions. The integration of structured reasoning and verification mechanisms represents a significant step toward improving the accuracy and interpretability of automated question answwering over tbaular data.

\section{Future Work}
While the proposed NL-to-SQL pipeline demonstrated significant improvements over baseline methods, an analysis of system outputs revealed several challenges affecting query accuracy and reliability. These errors primarily fall into two categories:
\newline 1. Complex Numerical Reasoning Errors: The system exhibited difficulties in handling queries that required multi-step numerical reasoning, particularly those involving ranking, aggregation, and filtering operations.
\newline 2. Categorical Misclassification: The system occasionally misclassified categorical values, selecting responses that were semantically related but incorrect.
% \newline Future improvements in large language models (LLMs) and NL-to-SQL frameworks can help address these challenges. 
\newline Advances in LLM architectures with better contextual understanding, improved token representations for tabular data, and stronger reasoning capabilities could enhance the accuracy of query generation. Additionally, developing more structured NL-to-SQL frameworks that incorporate explicit schema understanding, enhanced query verification, and iterative refinement processes may further improve performance. Combining these advancements with more effective post-processing techniques and adaptive learning strategies could lead to a more reliable NL-to-SQL system.

\section*{Acknowledgments}

We would like to thank the organisers of \emph{SemEval 2025 Task 8: Question Answering over Tabular Data} for providing the dataset as well as being extremely helpful for the entire duration of the task. We would also like to thank Shaz Furniturewala for his support and guidance throughout the process of conducting this research.

\bibliography{custom}
\newpage
\appendix
\section{Appendix}
\subsection{Prompt Templates}

\begin{tcolorbox}[
  colback=gray!5!white,
  colframe=gray!75!black,
  title=SQL Query Prompt Template,
  enhanced,
  breakable,
  sharp corners,
  listing only,
  listing options={
    basicstyle=\ttfamily\small,
    breaklines=true,
    breakatwhitespace=false,
    columns=fullflexible,
    keepspaces=true,
    showstringspaces=false,
    escapeinside={(*@}{@*)} % Allow LaTeX in the listing if needed
  }
]
You are a PostgreSQL expert. Given an input question, create a syntactically correct PostgreSQL query to run. Unless otherwise specified.

Here is the relevant table info: \{table\_info\}

Most columns have intuitive names.
Return the entire row(s) that contain the final answer in context of the original question based strictly on the SQL table you are given (always use SELECT (*)). Every question will be answered only from the table provided, no other source of data.

Below are a number of examples of questions and their corresponding PostgreSQL queries.
\end{tcolorbox}

\begin{tcolorbox}[
  colback=gray!5!white,
  colframe=gray!75!black,
  title=Final Response Prompt Template,
  enhanced,
  breakable,
  sharp corners,
  listing only,
  listing options={
    basicstyle=\ttfamily\small,
    breaklines=true,
    breakatwhitespace=false,
    columns=fullflexible,
    keepspaces=true,
    showstringspaces=false,
    escapeinside={(*@}{@*)} % Allow LaTeX in the listing if needed
  }
]
Given the following user question and row(s) containing the answer, infer and answer the user question in exactly the format expected. You are also given columns headers for the table from which the row is extracted for context.
Answer only the user question directly with the information from the SQL rows given to you. Answers should strictly contain only the value expected, NOTHING ELSE. Ensure you respond only with values directly from the rows, do not write full sentences. 
The following answer formats are expected based on the question asked:
Boolean: Valid answers include True/False. If a question expects a yes/no answer, respond strictly only with True or False.
Category: A value from a cell (or a substring of a cell) in the dataset.
Number: A numerical value from a cell in the dataset, which may represent a computed statistic (e.g., average, maximum, minimum).
List: A list containing a fixed number of categories or numbers. The expected format is: "['cat', 'dog']". 
Columns available in the dataset: \{column\_headers\}
Question: \{inputs['question']\}
SQL Result: \{inputs['result']\}
Answer:
\end{tcolorbox}

\end{document}